\documentclass{article}

\usepackage[final]{corl_2020} 

\title{PixL2R: Guiding Reinforcement Learning Using \\
Natural Language by Mapping Pixels to Rewards}

\usepackage{microtype}
\usepackage{graphicx}
\usepackage{subfigure}
\usepackage{booktabs} 

\usepackage{textcomp,xspace}
\newcommand\la{$\langle$}
\newcommand\ra{$\rangle$\xspace}

\usepackage{dsfont}
\usepackage{wrapfig}
\usepackage{caption}
\usepackage{color}
\usepackage{enumitem}

\author{Prasoon Goyal,
Scott Niekum,
Raymond J. Mooney \\
Department of Computer Science \\
University of Texas at Austin \\
\{pgoyal, sniekum, mooney\}@cs.utexas.edu
}

\begin{document}
\maketitle


\begin{abstract}
Reinforcement learning (RL), particularly in sparse reward settings, often requires prohibitively large numbers of interactions with the environment, thereby limiting its applicability to complex problems. To address this, several prior approaches have used natural language to guide the agent's exploration. However, these approaches typically operate on structured representations of the environment, and/or assume some structure in the natural language commands. In this work, we propose a model that directly maps pixels to rewards, given a free-form natural language description of the task, which can then be used for policy learning. Our experiments on the Meta-World robot manipulation domain show that  language-based rewards significantly improves the sample efficiency of policy learning, both in sparse and dense reward settings. 
\end{abstract}

\keywords{reinforcement learning, reward shaping, natural language} 


\section{Introduction}

Reinforcement learning (RL) problems often involve a trade-off between the ease of designing a reward function and the ease of learning from this reward. At one end of the spectrum, a sparse reward function -- e.g. a fixed positive reward for completing the task, and zero in all other states -- is easy to design, but does not give the learning agent any learning signal until it reaches the goal. As such, the agent requires considerable exploration before any learning can take place. At the other end of the spectrum, a dense reward function -- e.g. distance to the next waypoint -- can be specified to provide the agent with a stronger learning signal, but is often harder to design and tune compared to sparse reward functions. 
To get around the challenge of reward design, learning from demonstrations is a popular approach \cite{argall2009survey,gao2012survey}; however, providing demonstrations to robots requires teleoperating or kinesthetic teaching, which is difficult and time-consuming to provide, particularly for non-experts.
As such, several methods have been proposed recently, which involve guiding an agent using natural language commands which are quick and easy to provide \cite{luketina2019survey}.

While promising, these techniques are still quite restrictive, often requiring object properties to be predefined \cite{macglashan2014training,williams2018learning}, and/or assuming some structure in the natural language commands \cite{bahdanau2018learning}, which is challenging to scale.
Other techniques are applicable to only a restrictive set of environments, such as those with discrete action spaces \cite{goyal:ijcai19}. In this work, we propose a framework that makes no such assumptions, and directly learns to map pixels to rewards for continuous control given a free-form natural language description of the task. 

Our approach contains two phases -- 
(1) a supervised learning phase that takes in paired (trajectory, language) data and learns a model of relatedness between a trajectory and a language command, and (2) a policy training phase with a standard RL setup with an additional linguistic description of the task, wherein the relatedness model is used to generate intermediate rewards using the currently executed trajectory and task description. 

\begin{figure*}[t]
\centering
\subfigure{\includegraphics[height=1.2in]{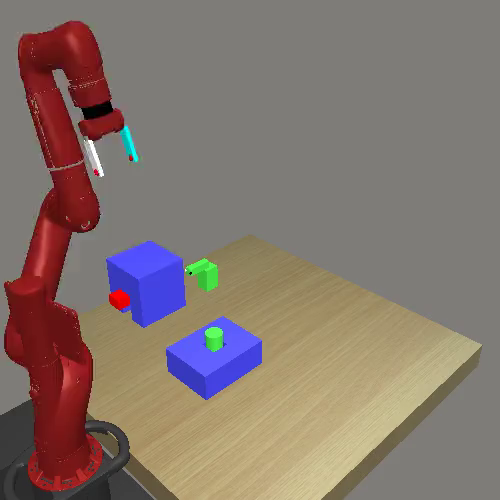}}
\subfigure{\includegraphics[height=1.2in]{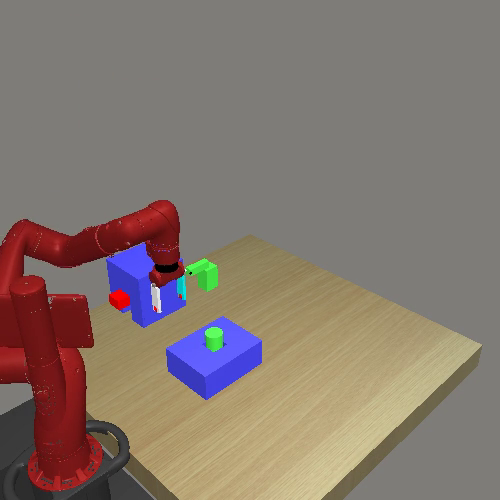}}
\subfigure{\includegraphics[height=1.2in]{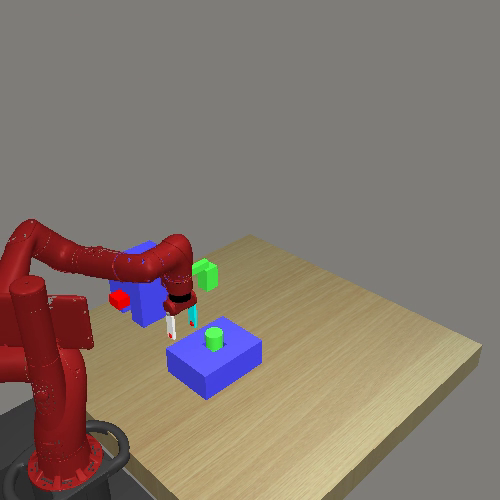}}
\subfigure{\includegraphics[height=1.2in]{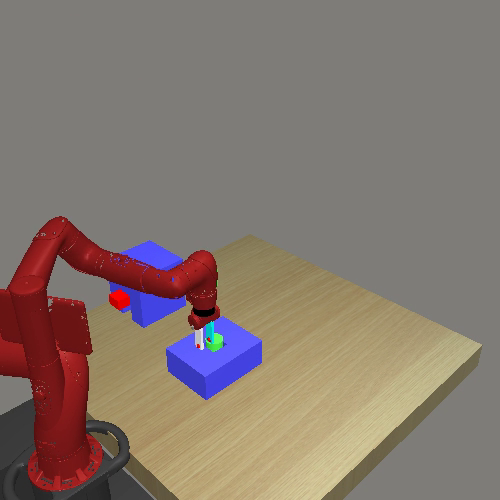}}
\caption{A simulated robot completing a task (``push the green button") in the Meta-World domain.}
\label{fig:metaworld}
\vspace{-10pt}
\end{figure*}

For instance, consider the domain shown in Figure~\ref{fig:metaworld}, which is adapted from the recently released Meta-World benchmark \cite{yu2019metaworld}. 
Here, we want the robot to press the green button. Different tasks in this domain require interacting with different objects.
In a sparse reward setting, the agent is given a non-zero reward only upon successfully interacting with the pre-selected object. In the absence of any other learning signal, the agent will explore randomly until it accidentally completes the desired task. Using natural language to describe the task and generating intermediate rewards from these descriptions can guide the agent towards the goal, significantly speeding up learning.

Our experiments on a diverse set of tasks in the Meta-World domain demonstrate that the proposed approach results in improved sample efficiency during policy learning, both in sparse and hand-designed dense reward settings. This motivates a new paradigm where language could be used to improve over hand-designed rewards, which may be suboptimal owing to the difficulty of designing rewards by hand.

\section{Related Work}

A number of prior approaches have been proposed to use language to guide a learning agent \cite{tellex2020robots}. 

Some approaches involve mapping natural language instructions directly to an action sequence to be executed.
Tellex et al. \cite{tellex2011understanding}
dynamically instantiate a graphical model given a language command, from which a plan for the agent is inferred.
Sung et al. \cite{sung2018robobarista}
learn a neural network to predict relatedness between \la trajectory, language\ra pairs and \la trajectory, point cloud\ra pairs, which is then used to find the most likely trajectory given a new language and point cloud.
Our approach is different from these approaches in that we use language to generate a reward for the current state, that can then be used to learn a policy using standard RL, which is a more general setting that does not require knowledge of the environment dynamics, and can also work in more complex environments because of the policy learning phase.

Several prior approaches map natural language to a reward function \cite{macglashan2014training,arumugam2017accurately,williams2018learning}, but assume a specific structure of the reward function, while our approach does not make any such assumptions.

Some approaches use a predefined set of linguistic instructions to guide the learning agent.
Kuhlmann et al. \cite{kuhlmann2004guiding} and Branavan et al. \cite{branavan2012learning} find the most relevant instruction to follow at each state.
Kaplan et al. \cite{kaplan2017beating} and Waytowich et al. \cite{waytowich2019grounding} learn a neural network that predicts the similarity between a natural language instruction and a state, and use that to follow a fixed sequence of natural language commands.
These approaches use hand-designed features, whereas we propose to learn the association between language and trajectories from a small set of human-provided descriptions.



Some approaches learn to ground language while interacting with the environment \cite{branavan2012learning2,misra2018mapping,bahdanau2018learning}.
Our approach involves a separate supervised learning phase to ground language, which does not require interacting with the environment.

Fu et al. \cite{fu2019language}
learn a language-conditioned reward function, but require knowledge of environment dynamics to compute the optimal policy during training.
Narasimhan et al. \cite{narasimhan2015language}
use natural language to transfer dynamics across environments.
Blukis et al. \cite{blukis2018mapping}
generate a state visitation distribution given a natural language instruction, which is then used to generate rewards for policy training.
Harrison et al. \cite{harrison2017guiding} learn a distribution of states and actions given a natural language command, which is used for policy shaping.
Lynch et al. \cite{lynch2020grounding} learn a goal-image- and language-conditioned policy, using behavior cloning.
Andreas el al. \cite{andreas2017learning} use natural language as a parameter space in supervised and reinforcement learning tasks.
Paxton et al. \cite{paxton2019prospection} generate subgoals from natural language descriptions of the task, which are then used by a controller to predict actions.
Goyal et al. \cite{goyal:ijcai19}
use a similar framework as us, but their approach uses only the actions to generate language-based rewards, without taking into account the states, and requires the action space to be discrete, which is not applicable to most robotics tasks.

Our setting is related to the problem of vision-language navigation (VLN) \cite{anderson2018vision}; while techniques in VLN work with complex multi-step instructions but predominantly discrete action spaces, the approach presented here is applicable to both discrete and continuous action spaces, but currently works with relatively simple instructions. Combining the strengths of both these settings to develop approaches that can work with complex instructions in continuous actions spaces is an interesting direction for future work.

\section{Approach}

Reinforcement learning consists of an agent interacting with an environment. The learning problem is typically represented using a Markov Decision Process (MDP) $M = \langle S, A, T, R, \gamma \rangle$. Here, $S$ is the set of all states in the environment, $A$ is the set of actions available to the agent, $T : S \times A \times S \rightarrow [0, 1]$ is the transition function of the environment, $R : S \times A \rightarrow \mathds{R}$ is the reward function, and $\gamma \in [0, 1]$ is a discount factor.

At timestep $t$, the agent observes a state $s_{t} \in S$, and takes an action $a_{t} \in A$, according to some policy $\pi : S \times A \rightarrow [0, 1]$. The environment transitions to a new state $s_{t+1} \sim T(s_{t}, a_{t}, \cdot)$, and the agent receives a reward $R_{t} = R(s_{t}, a_{t})$.
The goal is to learn a policy $\pi$, such that the expected future return, $G_{t} = \sum_{t=0}^{T} \gamma^{t} R_{t}$, is maximized.

\begin{wrapfigure}{r}{2.8in}
\vspace{-20pt}
\includegraphics[width=2.8in]{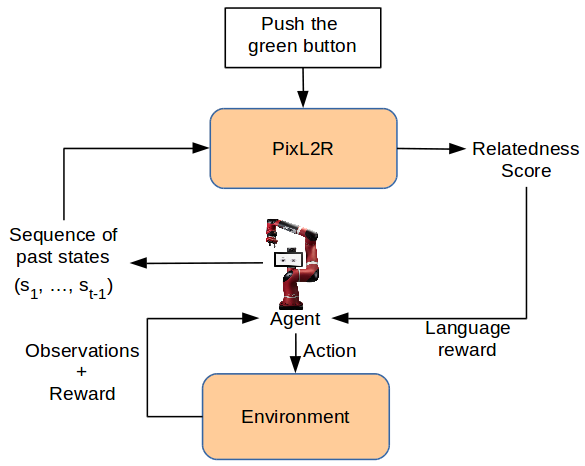}
\caption{Overview of the approach}
\label{fig:approach-overview}
\vspace{-10pt}
\end{wrapfigure}

In this work, we use an extension of the standard MDP, defined as $M' = \langle S, A, T, R, \gamma, L \rangle$, where $L$ is an instruction describing the task using natural language, and the other quantities are as defined above. We use the following two-phase framework for learning in an MDP with a natural language description of the task (Figure~\ref{fig:approach-overview}).

\textbf{Phase 1}: A neural network (PixL2R) is trained to predict whether a given trajectory and language are related or not. This requires paired \la trajectory, language\ra data in the environment. We describe this phase in detail in Section~\ref{sec:pixl2r}.

\textbf{Phase 2}: Next, a policy is trained for a new task -- in addition to the extrinsic reward from the environment, the agent additionally gets a language command describing the task. At every step, the agent's trajectory so far is compared against the description of the task using the trained PixL2R model and the relatedness scores predicted by the model are used to generate intermediate rewards for reward shaping \cite{ng1999policy}. Section~\ref{sec:policy-training} describes this phase.

Note that the trained PixL2R model can be used during policy learning for a wide variety of downstream tasks, insofar as the objects and linguistic vocabulary in these tasks closely match the data used to train the PixL2R model. Thus, the cost of training PixL2R is amortized across all the downstream tasks.



\subsection{PixL2R: Pixels and Language to Reward}
\label{sec:pixl2r}

First, a relatedness model -- PixL2R -- between a trajectory and a language is trained given paired data using supervised learning.

\subsubsection{Network Architecture}
\label{sec:arch}

The inputs to the network consist of a trajectory and a natural language description. Representing the trajectory using a single sequence of frames may be prone to perceptual aliasing and occlusion. 
Thus, our network architecture is designed to take multiple views as inputs. We use three different viewpoints in our experiments (see Figure~\ref{fig:viewpoints} in the Appendix), but it is straightforward to generalize to more or fewer viewpoints. 
In our ablation experiments, we compare the model described here with a model that takes a single viewpoint as input.

An independent CNN is used for encoding the sequence of frames from each viewpoint to generate a fixed size representation for each frame. These sequence of vectors are concatenated across the views to generate a single sequence of fixed size vectors, which is then passed through a two-layer LSTM to get an encoding of the entire trajectory. 

The language description is converted to a one-hot representation, and passed through an embedding layer, followed by a two-layer LSTM. 
The outputs of the LSTMs encoding the trajectory and the language are then concatenated, and passed through a sequence of fully-connected layers to generate a relatedness score. 
See Figure~\ref{fig:arch} in the Appendix for a diagram of the neural network.

\subsubsection{Data Augmentation}

\paragraph{Frame dropping.} After sampling a trajectory, each frame is independently selected with a probability of 0.1.
The resulting sequence of frames is passed through the network. This makes the training faster by reducing the input size, as well as making the network robust to minor variations in trajectories. During policy training, the trajectories are subsampled to keep 1 frame in every 10.

\paragraph{Partial trajectories.} Since during policy training the model will have to make predictions for partial trajectories, we use partial trajectories during supervised training as well. Given a trajectory of length $L$, we sample $l \sim Uniform\{1, \ldots, L\}$, and use the first $l$ frames of the trajectory.

\subsubsection{Training Objectives}
\label{sec:objectives}


\paragraph{Classification.} First, we trained the neural network using binary classification. The final output of the network is a two-dimensional vector, corresponding to the logits for the two classes -- \texttt{RELATED} and \texttt{UNRELATED}. The network is trained to minimize the cross-entropy loss.

As mentioned above, we train the model with partial trajectories of different lengths to better match the distribution of trajectories that will be seen during policy learning. However, partial trajectories might sometimes be hard to classify as related or unrelated to the description, since it requires extrapolating the complete path the agent will follow. Our preliminary experiments suggest that these harder to classify examples affect learning -- on unseen complete trajectories, a model trained with only complete trajectories has a lower error compared to a model trained on both complete and partial trajectories.
This motivated us to experiment with an alternative  regression setting described next.

\paragraph{Regression.}  In this setting, the model predicts a single relatedness score between the given trajectory and language, which is mapped to $[-1, 1]$ using the $\tanh()$ function. The ground truth score is defined as $s \cdot \frac{l}{L}$, where $s=1$ for \texttt{RELATED} and $s=-1$ for \texttt{UNRELATED} pairs, $l$ is the length of the incomplete trajectory and $L$ is the length of the complete trajectory as described above. Thus, given a description, a complete related trajectory has a ground truth score of $1$, while a complete unrelated trajectory has a score of $-1$. Shorter trajectories smoothly interpolate between these values, with very small trajectories having a score close to $0$. The network is trained to minimize the mean squared error. Intuitively, this results in a small loss when the model predicts the incorrect sign on short trajectories. As the trajectories become longer, incorrect sign predictions result in higher losses.

The network is trained end-to-end using an Adam optimizer \cite{kingma2014adam}.
We started by tuning the learning rate on a few different architectures -- of the 3 values we tried (1E-3, 1E-4, 1E-5), we found 1E-4 to work the best.
For the network architecture, we had 4 hyperparameters -- $D_1, D_2, D_3, D_4$ -- as shown in Figure~\ref{fig:arch}. For each of these hyperparameters, we searched over the following values -- \{64, 96, 128, 192, 256, 384, 512\}.
We experimented with 8 different combinations of values for the hyperparameters using random search, and selected the model with the best performance on the validation set. 
The source code and the data are available at \url{https://github.com/prasoongoyal/PixL2R}.

\subsection{Policy Learning Phase}
\label{sec:policy-training}

Having learned a PixL2R model as described above, the relatedness scores from the model can be used to generate language-based intermediate rewards during policy learning on new scenarios. During policy training, the agent receives a natural language description of the goal, in addition to the extrinsic reward from the environment. At every timestep, the PixL2R model is used to score trajectories executed by the agent against the given natural language description, to generate intermediate rewards. We used potential-based shaping rewards \cite{ng1999policy}, which are of the form $F(s_{t}) = \gamma \cdot \phi(s_{t}) - \phi(s_{t-1})$, where $s_{t}$ is the state at timestep $t$ and $\phi : S \rightarrow \mathds{R}$ is a potential function. In our case, $s_{t}$ is the sequence of states encountered by the agent up to timestep $t$ in the current episode. 
Ng et al. \cite{ng1999policy}
and 
Grzes et al. \cite{grzes2017reward}
show that potential-based shaping rewards do not change the optimal policy, i.e., the optimal policies under the original reward function $R$ and the new reward $R+F$ are identical.

For the classification setting, we used the potential function $\phi(s_{t}) = p_R(s_{t}) - p_U(s_{t})$, where $p_R$ and $p_U$ are the probabilities assigned by the model to the classes \texttt{RELATED} and \texttt{UNRELATED} respectively. For the regression setting, the relatedness score predicted by the model is directly used as the potential for the state. Note that for both the settings, the potential of any state lies in $[-1, 1]$.

\section{Domain and Dataset}

\vspace{-5pt}
\subsection{Description of the Domain}
\label{sec:domain}

We use Meta-World \cite{yu2019metaworld}, a recently proposed benchmark for meta-reinforcement learning, which consists of a simulated Sawyer robot and everyday objects such as a faucet, windows, coffee machine, etc. Tasks in this domain involve the robot interacting with these objects, such as turning the faucet clockwise, opening the window, pressing the button on the coffee machine, etc. Completing these tasks requires learning a policy for continuous control in a 4-dimensional space (3 dimensions for the end-effector position, and the fourth dimension for the force on the gripper). While the original task suite consists of only one object in every task, we create new environments which contain one or more objects in the scene, and the robot needs to interact with a pre-selected object amongst those. 
In a sparse reward setting, the agent is given a non-zero reward only on successfully interacting with the pre-selected object. In the absence of any other learning signal, the agent might have to learn to approach and interact with multiple objects in the scene in order to figure out the correct object. Using natural language to describe the task in addition to the sparse reward helps alleviate this issue.

\vspace{-5pt}
\subsection{Data Collection}
\label{sec:data}

First, 13 tasks were selected from the Meta-World task suite. This gave us a total of 9 objects to interact with (for 4 objects, multiple tasks can be defined, e.g. turning a faucet clockwise or counter-clockwise). We then created 100 scenarios for each task as follows: In each scenario, the task-relevant object is placed at a random location on the table. Then, a new random location is sampled, and one of the remaining objects is placed at this position. This process is repeated until the new random location is close to an already placed object. This results in 1300 scenarios in total, with a variable number of objects in each scenario. 

A policy was trained for each of these scenarios independently using PPO \cite{schulman2017proximal}, which was then used to generate one video of the robot completing the task in the scenario. For this purpose, we used the dense rewards defined in the original Meta-World benchmark for various tasks. The median length of trajectories across all generated videos is 131 frames. Note that our algorithm does not need the policies used to generate the videos, so they could also be collected using human demonstrations.

To collect English descriptions of these tasks, Amazon Mechanical Turk (AMT) was used. 
The workers were first provided with the instructions and an example trajectory with a possible description. They were then shown a video and were given 4 possible descriptions to choose from. Only workers that passed this basic test were allowed to provide descriptions for the main tasks.\footnote{The objects used for the example and the test are different from those used in the main tasks.} Each worker was asked to provide descriptions for 5 videos, which were sampled from the 1300 scenarios with the constraint that no two videos in the selected videos belong to the same task. We used simple heuristics (such as number of words and characters in the descriptions) to automatically filter out clearly bad descriptions. 


Interestingly, most of the descriptions involve only the object being manipulated, with no reference to other objects in the scene. As such, a description collected for one scenario for a task can be paired with any of the 100 scenarios for the corresponding task. Therefore, we collected a total of 520 descriptions, which gives us 40 descriptions per task on average.

For each task, 79 scenarios were used for training, 18 for validation, and 3 for testing. Similarly, the descriptions for each task were split as follows -- 5 for validation, 3 for testing, and the remaining for training (since there could be variable number of descriptions per task).

\begin{wrapfigure}{r}{2.8in}
\vspace{-5pt}
\includegraphics[width=2.8in]{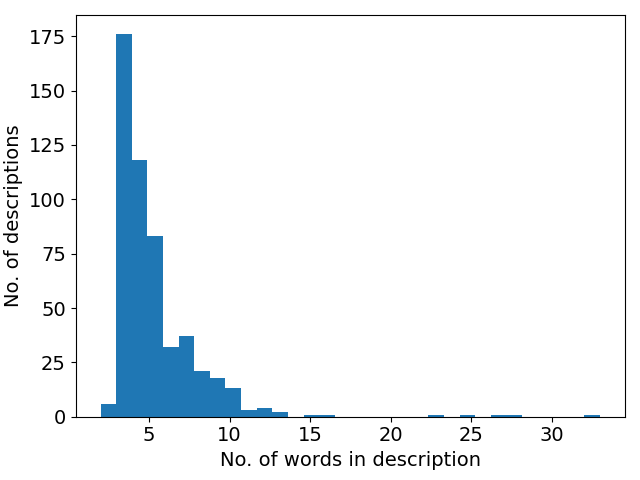}
\vspace{-15pt}
\caption{Distribution of number of words per description in the collected dataset.}
\label{fig:descr-lens}
\vspace{-15pt}
\end{wrapfigure}

The distribution of number of words per description is shown in Figure~\ref{fig:descr-lens}.
Our dataset contains 264 unique words, while our training set contains 248 unique words. See the Appendix (Section~\ref{sec:data-details}) for sample descriptions and more details about the data collection process.

Given pairs of related \la trajectory, language\ra, positive examples were generated by pairing a scenario for one of the 13 tasks with a randomly sampled description of the corresponding task. To generate negative examples, if a scenario contains more than one object, then it was paired with the description of the task corresponding to one of the alternate objects in the scene; if there was only one object in the scene, then it was paired with the description of any of the remaining 12 tasks. Using such a scheme for generating negative examples is important because naively creating pairs of trajectories with descriptions of any other task randomly might result in most negative examples lacking the task-relevant object mentioned in the description. As such, the network might learn to use the \textit{presence} of the mentioned object to compute relatedness, instead of whether the mentioned object is being \textit{interacted with}. 

\section{Experiments}

\subsection{Policy Training with Language-based Rewards}

To empirically evaluate the effectiveness of PixL2R, the following setup was used. For each of the 13 tasks, a policy was trained for the 3 test scenarios using the PPO algorithm. Each policy training was run for 500,000 timesteps, and the number of successful completions of the task were recorded. The maximum episode length was restricted to 500 timesteps. The robot's end-effector was set to a random position within a predefined region at the beginning of each episode.

First, policy training was run with 15 random seeds, both in the sparse reward setting  (\texttt{Sparse}; 1 if the agent reaches the goal, and 0 otherwise) and the hand-designed dense reward setting (\texttt{Dense}; defined in the original Meta-World benchmark). Then, a Kruskal-Wallis test was used for each scenario to identify scenarios where the number of successful episodes using dense rewards was statistically significantly more than the number of successful episodes using sparse rewards.
All subsequent comparisons were done on the 16 (out of 39) scenarios for which this was true. 
Intuitively, these 16 tasks are too difficult to learn from sparse rewards, while they can be learned using dense rewards. Therefore, language-based dense rewards should be useful on these tasks. The remaining tasks are presumably either too simple that they can be learned with sparse rewards alone, or are too difficult to learn within 500,000 timesteps even with hand-designed dense rewards.

Then, for each of the 16 selected scenarios, a policy was trained with language-based rewards using the regression setting, in addition to the sparse rewards  (\texttt{Sparse+RGR}).
For each scenario, 5 policies were trained with different seeds for each of the 3 test descriptions, resulting in a total of 15 policy training runs per scenario.

\begin{figure*}[t]
\centering
\subfigure{\includegraphics[width=0.48\textwidth]{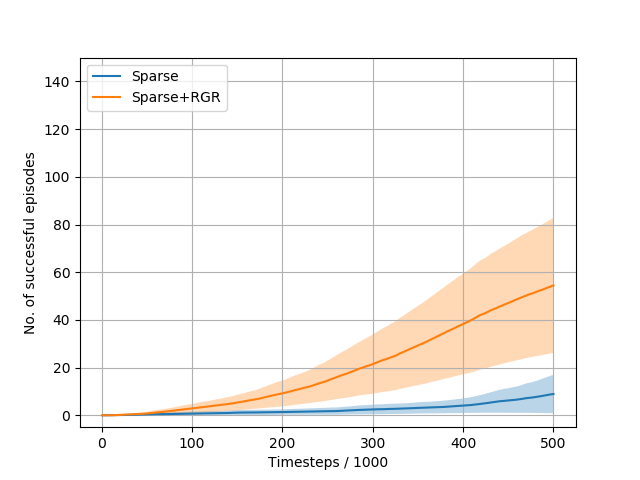}}
\subfigure{\includegraphics[width=0.48\textwidth]{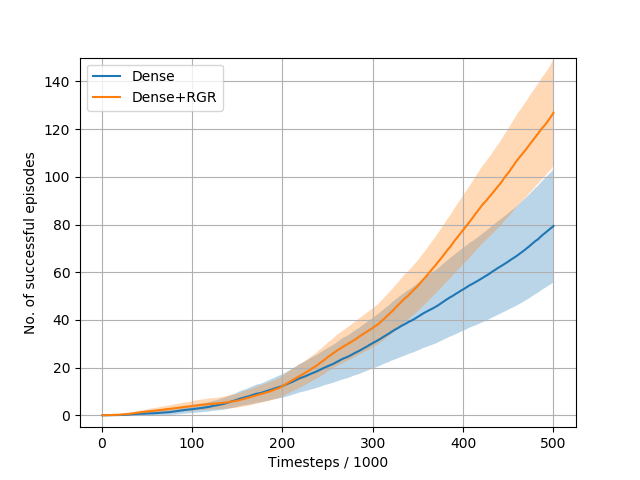}}
\vspace{-10pt}
\caption{A comparison of policy training curves for different reward models. The shaded regions denote 95\% confidence intervals.}
\vspace{-10pt}
\label{fig:learning-curves}
\end{figure*}

A comparison of policy training curves for \texttt{Sparse} and \texttt{Sparse+RGR} rewards is shown in Figure~\ref{fig:learning-curves} (left). Each curve is obtained by averaging over all runs (16 scenarios $\times$ 15 runs per scenario) for that reward type. 
The results verify that using language-based rewards in addition to sparse rewards result in higher performance on average than using only sparse ones. 

Next, language-based rewards were used in addition to hand-designed rewards using a similar methodology, and the corresponding learning curves for \texttt{Dense} and \texttt{Dense+RGR} are shown in Figure~\ref{fig:learning-curves} (right). Interestingly, we find that using language-based rewards in conjunction with hand-designed rewards result in an improvement even over hand-designed rewards. A plausible explanation is that the hand-designed dense rewards in Meta-World are suboptimal, since the reward function for each task consists of parameters that require tuning, highlighting the complexity of reward design mentioned in the introduction.
This result motivates a novel paradigm wherein coarse dense rewards could be designed by hand, and the proposed framework can be used to get a further improvement in policy training efficiency by using natural language.

Further, the statistical significance was computed to compare the reward functions.
For each type of reward, first the average number of successful episodes was computed across all the 15 runs for each scenario, giving 16 mean successful episode scores per reward type. 
Since the number of successful episodes across different scenarios vary quite a bit, the mean scores for each scenario were scaled to be at most 1, by dividing by the maximum value of the mean score across all reward types for that scenario (including the reward types used in ablation experiments described in Section~\ref{sec:ablation}).

A Wilcoxon signed-rank test was then performed between the sets of normalized scores across reward types.
\texttt{Sparse+RGR} was found to be statistically significantly better than \texttt{Sparse} (p-value=0.007) and \texttt{Dense+RGR} was found to be statistically significantly better than \texttt{Dense} (p-value=0.034) rewards, at a 5\% significance level .
Thus, the proposed approach can be used to make policy learning more sample efficient in both sparse and dense reward settings.

Having established that policy learning works better with the language-based rewards, we ran ablation experiments (described below) and analyzed the supervised learning phase (see Section~\ref{sec:expts-details} of the Appendix) to better understand our design choices and to inspect what factors most affect the efficiency of policy learning.

\subsection{Ablations}
\label{sec:ablation}


All the ablation experiments were performed with language-based rewards added to dense rewards, since most applications of RL in robotics currently use dense hand-designed rewards (which could be suboptimal for complex tasks).

\begin{wraptable}{r}{2.5in}
\vspace{-10pt}

\begin{tabular}{lrr}
\hline
\multicolumn{1}{c}{\textbf{Setting}} & \multicolumn{1}{c}{\textbf{\begin{tabular}[c]{@{}c@{}}Mean\\ Successful\\ Episodes\end{tabular}}} & \multicolumn{1}{c}{\textbf{\begin{tabular}[c]{@{}c@{}}p-value\\ w.r.t.\\ Dense\end{tabular}}} \\ \hline 
Dense                          & 79.4               & -                     \\ 
Dense+RGR                      & 126.9              & 0.0340                \\ \hline 
LastFrame                      & 133.5              & 0.0114                \\ 
MeanpoolLang                   & 138.3              & 0.0004                \\ 
MeanpoolTraj                   & 78.4               & 0.9601                \\ 
SingleView                     & 100.4              & 0.3789                \\ 
Dense+CLS                      & 102.0              & 0.6384                \\ \hline
\end{tabular}
\captionof{table}{Comparison of various ablations to the Dense+RGR model.}
\label{tbl:ablations}

\vspace{-10pt}
\end{wraptable}

    (1) \textbf{LastFrame}: To analyze whether using the full sequence of frames contains more information than the last frame, instead of using the sequence of frames in the trajectory, only the last frame of the trajectory was used, both for training the PixL2R model, as well as for policy training. \\
    (2) \textbf{MeanpoolLang}: To study if the temporal ordering of the words in the description is useful, the LSTM used to encode the language was replaced with the mean-pooling operation. \\
    (3) \textbf{MeanpoolTraj}: To study if the temporal ordering of the frames in the trajectory was useful, the LSTM used to encode the sequence of frames was replaced with the mean-pooling operation. \\
    (4) \textbf{SingleView}: To study the impact of perceptual aliasing and/or occlusion when using a single viewpoint, instead of using 3 viewpoints for the trajectory, only 1 viewpoint was used. A model was trained with \emph{each} of the three viewpoints in the supervised learning phase, and the model with the best validation score was used for policy learning. \\
    (5) \textbf{Dense+CLS}: Instead of the regression loss, classification loss was used, to understand the benefit of using regression loss when working with partial trajectories.


For each ablation, the same setup was used as for \texttt{Dense+RGR}.
This model is used to generate rewards for policy training, for each of the 16 scenarios with 5 random seeds for all the 3 descriptions as before.
The mean successful episodes across all runs are reported in Table~\ref{tbl:ablations}. Further, the p-values for Wilcoxon tests between each ablation and the \texttt{Dense} rewards is reported, from which we can make the following observations:
\begin{itemize}[leftmargin=10pt] \itemsep0em
    \item Using only the last frame (\texttt{LastFrame}), or using mean-pooling instead of an LSTM to encode the language (\texttt{MeanpoolLang}) does not substantially affect policy learning efficiency. In both these cases, the resulting model is still statistically significantly better than \texttt{Dense} rewards. Both of these results agree with intuition, since the progress in the task can be predicted using the last frame alone, and since the linguistic descriptions are not particularly complex in the given domain, simply looking at which words are present or absent is often sufficient to identify the task without using the ordering information between the words. 
    \item Using mean-pooling instead of an LSTM to encode the sequence of frames (\texttt{MeanpoolTraj}) drastically reduces the number of successful episodes, and results in no statistically significant improvement over \texttt{Dense}. Again, this agrees with intuition, since it is not possible to infer the direction of movement of the robot from an unordered set of frames.
    \item Using a single view instead of multiple views (\texttt{SingleView}) results in a smaller increase in the number of successful episodes, which is no longer statistically significant over \texttt{Dense}.
    As mentioned earlier, using frames to represent trajectories requires addressing challenges such as perceptual aliasing and occlusion, and these ablation results suggest that using multiple viewpoints alleviates these issues.
    \item Using classification loss instead of regression (\texttt{Dense+CLS}) also leads to a drop in performance, again making the resulting improvements no longer statistically significant. This is consistent with our initial observation (Section~\ref{sec:objectives}), wherein, the learning problem becomes more difficult due to partial trajectories when the classification loss is used.
\end{itemize}

It is worth noting that while these ablations agree with intuition, and therefore suggest that the model is extracting meaningful information from trajectories and language descriptions, the performance of these variants depends crucially on the domain. For instance, an environment that is not fully observable in the last frame might show a significant drop in performance when using only the last frame instead of the full trajectory.
\section{Conclusion}

We proposed an approach for mapping pixels to rewards, conditioned on a free-form natural language description of the task. Given paired \la trajectory, language\ra data, first, a relatedness model -- PixL2R --  is learned between a sequence of states and a natural language description using supervised learning. This model is then used to generate intermediate rewards for policy learning using a natural language task description. 
Our experiments on a simulated robot manipulation domain show that the proposed approach can significantly speed up policy learning, both in sparse and dense reward settings. The proposed technique can be used in a novel RL training paradigm, wherein language-based rewards can be used to make training efficient over coarse hand-designed dense rewards.


The proposed approach can be extended in multiple ways. 
First, further experimentation on a richer domain could be used to analyze generalization to new tasks with novel compositions of objects and actions seen during training. 
Secondly, the current model only works for a single instruction and could be extended to use a sequence of instructions, for instance, by starting with the first instruction in the sequence, and transitioning to the next instruction when the prediction of the PixL2R model is above a threshold. 
Next, PixL2R currently encodes the trajectory and language independently, which are then concatenated to obtain a relatedness score. For more complex domains, it might be helpful to use an attention-based model to learn a mapping between spatio-temporal regions of the trajectory and words or phrases in the language.
Finally, it may be useful to fine-tune the PixL2R model on trajectories seen during policy learning.

\newpage

\section*{Acknowledgements}

This work has taken place in the Personal Autonomous Robotics Lab (PeARL) at The University of Texas at Austin. PeARL research is supported in part by the NSF (IIS-1925082, IIS-1724157, IIS-1638107, IIS-1749204), ONR (N00014-18-2243), and an Amazon Research Award. This research was also sponsored by the Army Research Office and was accomplished under Cooperative Agreement Number W911NF-19-2-0333. The views and conclusions contained in this document are those of the authors and should not be interpreted as representing the official policies, either expressed or implied, of the Army Research Office or the U.S. Government. The U.S. Government is authorized to reproduce and distribute reprints for Government purposes notwithstanding any copyright notation herein.


\nocite{kroemer2019review}
\bibliography{main}  

\newpage

\appendix
\section{Approach Details}
\label{sec:approach-details}

\begin{figure}[h]
\centering
\subfigure{\includegraphics[width=1in]{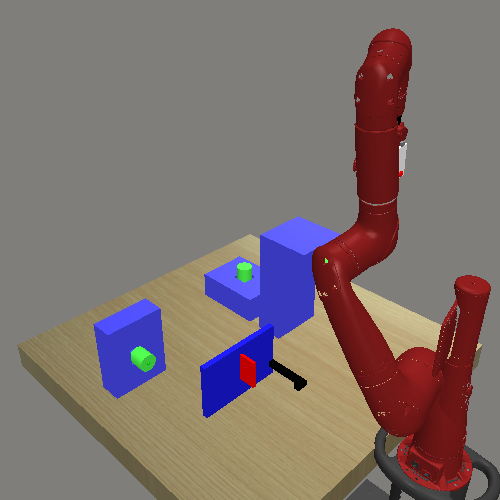}}
\hspace{1mm}
\subfigure{\includegraphics[width=1in]{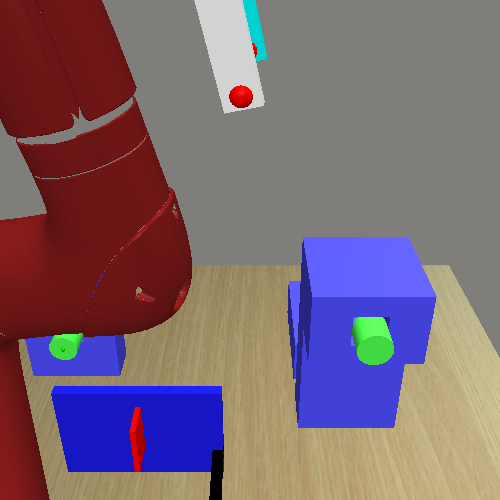}}
\hspace{1mm}
\subfigure{\includegraphics[width=1in]{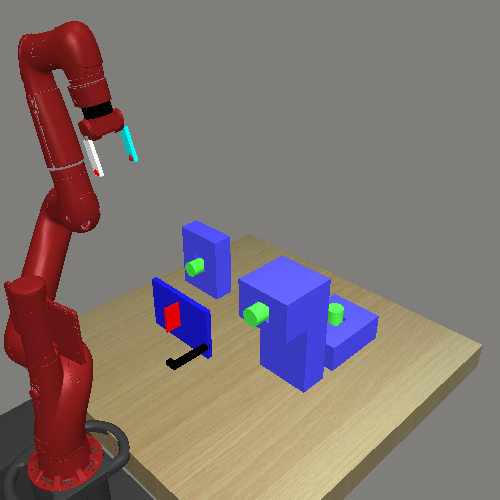}}
\caption{Viewpoints used for data collection and experiments.}
\label{fig:viewpoints}
\end{figure}
\begin{figure*}[h]
\centering
\includegraphics[width=\textwidth]{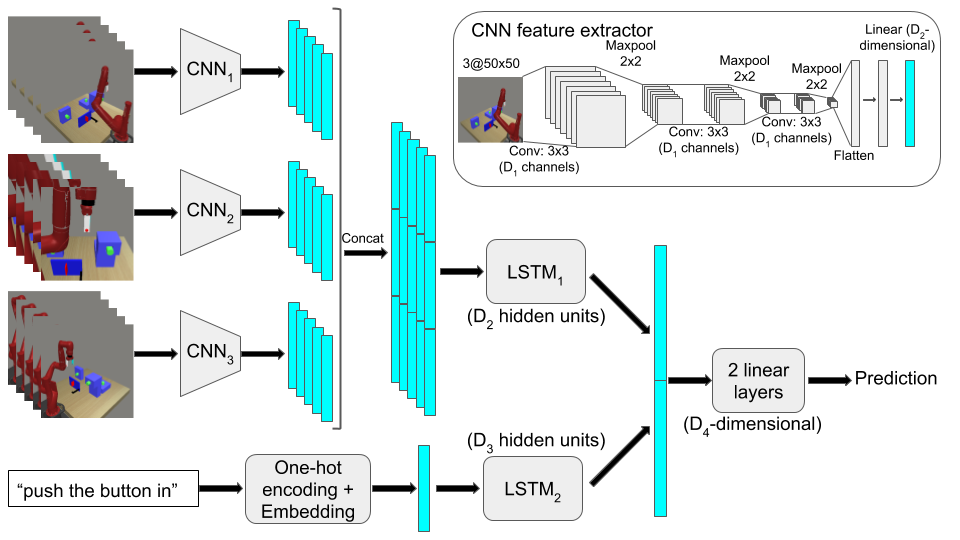}
\vspace{-10pt}
\caption{Neural network architecture: The sequence of frames from the three viewpoints are passed through three separate CNN feature extractors. The resulting feature vectors are concatenated across views. The sequence is then passed through an LSTM to obtain an encoding of the trajectory. The given linguistic description is converted to one-hot representation, and passed through an embedding layer, followed by an LSTM. The outputs of the two LSTMs is concatenated and passed through a sequence of 2 linear layers (with a ReLU activation between them) to generate the final prediction.}
\label{fig:arch}
\end{figure*}

Figure~\ref{fig:viewpoints} shows the viewpoints used in our experiments. Figure~\ref{fig:arch} shows a diagram of the neural network architecture described in Section~\ref{sec:arch}.

\section{Data Collection Details}
\label{sec:data-details}

\begin{figure}[t]
\centering
\includegraphics[width=1.5in]{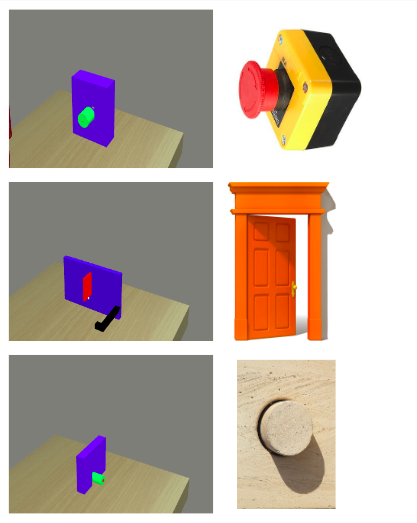}
\includegraphics[width=1.5in]{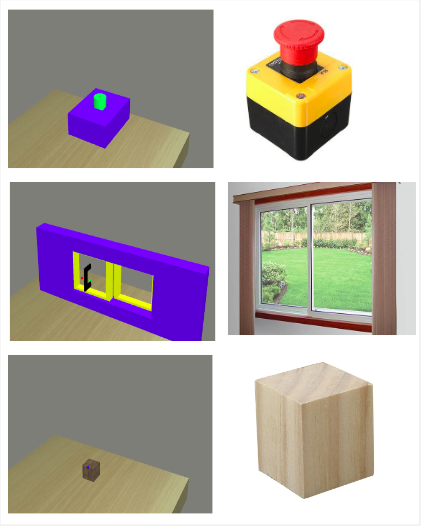}
\includegraphics[width=1.5in]{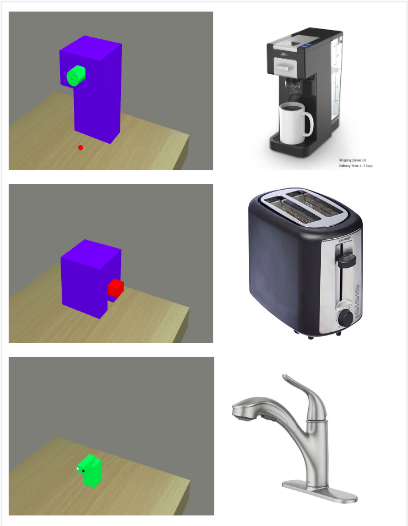}
\captionof{figure}{List of objects used}
\label{fig:amt-table}
\end{figure}

Since the models of the objects in the environment are coarse, it is usually non-trivial to recognize the real-world objects they represent from the models alone. To guide the AMT workers to use the names of real-world objects the models represent, we showed a table of the models with prototypical images of real-world objects that closely match the models (shown in Figure~\ref{fig:amt-table}). This enabled us to get descriptions that use the real-world object names, without priming the workers with specific words.\footnote{Despite using this technique, we still got some responses where people described the models directly instead of using the object names, e.g. ``Pull the red box out slightly in blue square." instead of using the word \textit{toaster}.}

Some examples of descriptions collected are shown in Table~\ref{tbl:descr-examples}.

\begin{table}[]
\centering
\begin{tabular}{cl}
\hline
\begin{tabular}[c]{@{}l@{}} 
Task Id\end{tabular} & Description                    \\ \hline 
0                   & Press the button.              \\
0                   & Pressing the button            \\
1                   & Push peg in to hole.           \\
1                   & Push the green button.         \\
2                   & Turn on the coffee maker       \\
2                   & push in the green button       \\
3                   & Push toaster handle down       \\
3                   & Push down the red block.       \\
4                   & pressing down the object       \\
4                   & pull down the red switch       \\
5                   & move the plate down            \\
5                   & push down the slider           \\
6                   & Close the door                 \\
6                   & Open the door.                 \\
7                   & twisting the cube              \\
7                   & rotate the object              \\
8                   & Rotate the lever anticlockwise \\
8                   & Turn the faucet to the right.  \\
9                   & rotating the object            \\
9                   & turn on the faucet             \\
10                  & Open the window.               \\
10                  & Open the yellow window.        \\
11                  & Slide the window to the left.  \\
11                  & Close the Window.              \\
12                  & pull out the green block       \\
12                  & Pull out the green piece       \\ \hline
\end{tabular}
\captionof{table}{Examples of descriptions collected using AMT.}
\label{tbl:descr-examples}
\end{table}

\section{Additional Experiments: Word-level Analysis}
\label{sec:expts-details}

In order to understand how the supervised learning phase is using different words in the description, the supervised model was used to make predictions on the test set, and the gradient of the loss was computed with respect to the continuous representation of the words in the descriptions (i.e. after the embedding layer). 
The mean of the absolute values of these gradients is then a measure of how much the prediction is affected by the corresponding word.
The values are reported in Table~\ref{tbl:word-analysis}, which were scaled so that the maximum value for any description is 1.

\begin{table}[]
\centering
\begin{tabular}{|r|rrrrrrr|}
\hline
\multicolumn{1}{|l|}{\textbf{}} & \multicolumn{7}{c|}{\textbf{Descriptions}}              \\ 
\multicolumn{1}{|c|}{\textbf{}} & \multicolumn{7}{c|}{\textbf{Average magnitude of gradient for each word}}       \\ \hline
1.                              & push  & the  & green  & button &       &      &         \\ 
                                & 0.53  & 0.30 & 1.00   & 0.94   &       &      &         \\ \hline
2.                              & push  & down & the    & red    & block &      &         \\ 
                                & 0.42  & 0.57 & 0.34   & 1.00   & 0.91  &      &         \\ \hline
3.                              & pull  & down & the    & lever  & on    & the  & toaster \\ 
                                & 0.16  & 0.31 & 0.15   & 0.75   & 0.58  & 0.36 & 1.00    \\ \hline
4.                              & turn  & on   & the    & faucet &       &      &         \\ 
                                & 0.94  & 1.00 & 0.44   & 0.87   &       &      &         \\ \hline
5.                              & slide & the  & green  & lever  & to    & the  & left    \\ 
                                & 0.52  & 0.23 & 0.94   & 1.00   & 0.77  & 0.30 & 0.78    \\ \hline
6.                              & open  & the  & window &        &       &      &         \\ 
                                & 0.83  & 0.32 & 1.00   &        &       &      &         \\ \hline
\end{tabular}
\caption{Average magnitude of gradients for different words in a description for the relatedness score prediction.}
\label{tbl:word-analysis}
\end{table}

First, we observe that for all the descriptions, the words describing the main object have a very high average gradient magnitude -- \textit{green} and \textit{button} in description 1, \textit{red} and \textit{block} in description 2, \textit{lever} and \textit{toaster} in description 3, \textit{faucet} in description 4, \textit{green} and \textit{lever} in description 5, and \textit{window} in description 6. 
Several verbs also have a high average gradient magnitude -- \textit{turn on} in description 4 and \textit{open} in window. Verbs in other descriptions do not have a high gradient magnitude because for those descriptions, the object affords only one possible interaction, thus making the verb less discriminatory. For the objects \textit{faucet} and \textit{window}, there are two possible actions each (\textit{turning the faucet on or off} and \textit{opening or closing the window}); thus the verb also carries useful information for these objects.

This analysis suggests that the model learns to identify the most salient words in the description that are useful to predict the relatedness between a trajectory and language.

\end{document}